\documentclass{article}

\usepackage{times}
\usepackage{graphicx} 
\usepackage{subfigure} 

\usepackage{natbib}

\usepackage{algorithm}
\usepackage{algorithmic}

\usepackage{amsmath,amsbsy}
\usepackage{amssymb}

\newtheorem{thm}{\rm\bf Theorem}
\newtheorem{lem}[thm]{Lemma}

\newcommand*\Bell{\ensuremath{\boldsymbol\ell}}

\newcommand{\dsum}{\displaystyle\sum}
\newcommand{\mbx}{\mathbf{x}}
\newcommand{\mbw}{\mathbf{w}}
\newcommand{\mbz}{\mathbf{z}}

\newcommand{\mba}{\mathbf{a}}
\newcommand{\mbb}{\mathbf{b}}

\newcommand{\dmax}{\displaystyle\max}

\usepackage{hyperref}

\usepackage[accepted]{icml2017}

\begin{document} 

\twocolumn[
\icmltitle{From Deep to Shallow: Transformations of Deep Rectifier Networks}

\icmlauthor{Senjian An}{senjian.an@uwa.edu.au}
\icmladdress{The University of Western Australia}
\icmlauthor{Farid Boussaid}{farid.boussaid@uwa.edu.au}
\icmladdress{The University of Western Australia}
\icmlauthor{Mohammed Bennamoun}{mohammed.bennamoun@uwa.edu.au}
\icmladdress{The University of Western Australia}
\icmlauthor{Jiankun Hu}{J.Hu@adfa.edu.au}
\icmladdress{The University of New South Wales}

\icmlkeywords{boring formatting information, machine learning, ICML}

\vskip 0.3in
]

\begin{abstract} 

In this paper, we introduce transformations of deep rectifier networks, enabling the conversion of deep rectifier networks into  shallow rectifier networks. We subsequently prove that any rectifier net of any depth can be represented by a maximum of a number of functions that can be realized by a shallow network with a single hidden layer. The transformations of both deep rectifier nets and deep residual nets are conducted to demonstrate the advantages of the residual nets over the conventional neural nets and the advantages of the deep neural nets over the shallow neural nets. In summary, for two rectifier nets with different depths but with same total number of hidden units, the corresponding single hidden layer representation of the deeper net is much more complex than the corresponding single hidden representation of the shallower net.  Similarly, for a residual net and a conventional rectifier net with the same structure except for the skip connections in the residual net, the corresponding single hidden layer representation of the residual net is much more complex than the corresponding single hidden layer representation of the conventional net. 

\end{abstract}

\section{Introduction}

 The application of deep learning networks to computer vision has resulted in remarkable successes in recent years. State-of-art performance has been achieved in a wide range of tasks such as  handwritten digit recognition \citep{ciresan2012multi}, object detection \cite{ren2015faster} and image classification \citep{krizhevsky2012imagenet,he2016deep,he2016identity}. Many of the deep learning architecture's characteristics such as the depth of the learning net, the skip connections in residual nets or the convolution in convolutional neural nets are believed to contribute to the successes of deep learning methods.  Although many theoretical works have been conducted to explain the success of deep learning networks, the justification remains challenging due to the lack of explicit relationships between the representations of classifiers under various deep learning architectures. For instance, to demonstrate the advantages of deep networks over shallow networks, one would first need to address the following two fundamental questions: Can the models in deep networks be represented by shallow networks? If so, what are their relationships? In this paper, we address these two questions by investigating the transformations of deep rectifier networks (where rectifier $\max(0,x)$ is the activation function in the nodes. To this end, we generalize conventional networks, for which each dimension of the output is a linear unit in the output layer, to a more general case where each dimension of the output is the maximum of a number of linear units. Such nets are termed as max-rectifier nets where the hidden nodes are activated by rectifiers while the output nodes are activated by a $\max$ operation on a number of linear units. We prove that the models of any deep rectifier network can be represented by a shallow max-rectifier net with a single hidden layer. We will analyse the advantages of deep nets over shallow nets, and the advantages of residual nets over conventional neural nets by using the number of the linear units in the max output layer and the number of hidden nodes of their corresponding shallow nets with a single hidden layer.

The main contributions of this paper include:  {\bf i)} the first development of transformations which convert deep rectifier nets into shallow nets; {\bf ii)} the analysis of the superior expressive power of deep rectifier nets from their explicit relationship to shallow nets; and {\bf iii)} the analysis of the superior power of deep residual nets over conventional rectifier nets without skip connections. The conventional rectifier nets will be referred to as plain nets hereafter, following \cite{he2016deep}.

{\bf Notations.} Throughout the paper, we use capital letters to denote matrices, lower case letters for scalar terms, and bold lower letters for vectors. For instance, we use $\mbw_i$ to denote the $i^{th}$ column of a matrix $W$, and use $b_i$ to denote the $i^{th}$ element of a vector $\mathbf{b}$.  For any integer $m$, we use $[m]$ to denote the integer set from 1 to $m$, i.e., $[m]\triangleq \{1,2,\cdots,m\}$. We use $I$ to denote the identity matrix with proper dimensions, $\mathbf{0}$ to denote a vector with all elements being 0, and $\mathbf{1}$ to denote a vector with all elements being 1.  $W\succeq 0$ and $\mathbf{b}\succeq 0$ denote that all elements of $W$ and $\mathbf{b}$ are non-negative while $W\preceq 0$ and $\mathbf{b}\preceq 0$ denote that all elements of $W$ and $\mathbf{b}$ are non-positive.

{\bf Organization.} The rest of this paper is organised as follows. Section 2 addresses the related work. Section 3 introduces the max-rectifier networks and investigate the transformations of plain rectifier networks. Section 4 considers the transformations of deep rectifier networks with full skip connections while Section 5 addresses the residual nets in particular. Section 6 compares the transformations of plain nets and residual nets. Finally Section 7 concludes the paper.

\section{Related Work}

The depth of neural networks has been investigated extensively in recent years to explain the superior expressive power of deep neural nets against shallow nets.  \citet{delalleau2011shallow} showed that the deep network representation of a certain family of polynomials can be much more compact (i.e., with less number of hidden units) than that provided by a shallow network. Similarly, with the same number of hidden units, deep networks are able to separate their input space into many more regions of linearity than their shallow counterparts \citep{Pascanu+et+al-ICLR2014b,montufar2014number}. 
\cite{eldan2015power} showed that there exists a simple  function on high dimensional space  expressible by a small 3-layer feedforward neural network, which cannot be approximated by any 2-layer network, to more than a certain constant accuracy, unless its width is exponentially increasing with the dimension of the data. \cite{cohen2015expressive} proved that besides a negligible set, all functions that can be implemented by a deep network of polynomial number of units, require exponentially large number of units in order to be realized (or even approximated) by a shallow network. 
\cite{mhaskar2016learning} demonstrated that deep (hierarchical) networks can approximate the class of compositional functions with the same accuracy as shallow networks but with exponentially lower number of training parameters as well as VC-dimension, while the universal approximation property holds both for hierarchical and shallow networks. 

The superior expressive power of deep residual nets
was analysed by \cite{veit2016residual} who showed that residual nets can be understood as a collection of many paths of various lengths and these paths enable the training of very deep networks by leveraging short paths. Unlike plain neural nets, paths through residual networks vary in length. 

All these aforementioned works address the compactness of the representations of functions through deep neural networks. The explicit relationship between deep and shallow representations was not addressed. In this paper, we establish the explicit relationship of a function's representations by deep rectifier networks and shallow networks, and this explicit relationship enables one to compare networks with different architectures and analyse the advantages of depth and skip connections.

\section{Transformations of Deep Rectifier Nets}

Consider a model of a rectifier net with $m$ hidden layers
\begin{equation}
\begin{array}{ll}
({\bf \mathsf{DrNet}})& 
\left\{\begin{array}{rcl}
f(\mbx)&=&c+\mba_0^T\mbx+\dsum_{j=1}^m\mba_k^T\max(0,\mbz_k)\\
\mbz_1 &=& W_1\mbx+\mbb_1\\
\mathbf{z}_i &=& W_i\max(0,\mbz_{i-1}), i\geq 2 \\
\end{array}\right.
\end{array}
\label{DrNet}
\end{equation}

In this model, the output of the $i^{th}$ layer is $\max(0,\mbz_i)\in\mathbb{R}^{l_i}$ where $l_i$ denotes the number of units of the $i^{th}$ layer. The input $\mbx$ is treated as the output of the $0^{th}$ layer and its dimension is denote by $l_0$. Correspondingly, $W_{i}\in\mathbb{R}^{l_i\times l_{i-1}}$ and $\mbb_i\in\mathbb{R}^{l_i}$ denote the weight matrix and the bias vector of the linear units from the $(i-1)^{th}$ layer to the $i^{th}$ layer. The outputs of hidden layers can be viewed as generated nonlinear features for nonlinear function representation. In traditional rectifier network, the final output is usually a linear function of the output of the last hidden layer where $\mba_k=0$ for $k<m$. 

For more compact representation of the deep rectifier net, let us denote   
\begin{equation}
\begin{array}{rcl}
W&=& \left[\begin{array}{ccccc}\mbb_1& W_{1} &  & & \\ \mbb_2&& W_{2}  & &\\
\vdots&&&\ddots&\\
\mbb_m&&&&W_{m}\\
\end{array}\right]
\end{array}
\label{Wmatrix}
\end{equation}
and
\begin{equation}
\begin{array}{rcl}
\overline{\mbz}_k &=& \left[\begin{array}{c} \mbz_1\\ \mbz_2 \\ \vdots\\\mbz_k
\end{array}\right], k=1,2,\cdots,m.
\end{array} 
\label{overlineZ} 
\end{equation}
Then the deep rectifier net, defined in (\ref{DrNet}), can be rewritten as
\begin{equation}
\begin{array}{ll}
({\bf \mathrm{DrNet}})& 
\left\{\begin{array}{rcl}
f(\mbx) & = & c+\mba_0^T\mbx+\dsum_{k=1}^m\mba_k^T\max(0,\mbz_k)\\
\overline{\mbz}_m &=& W\left[\begin{array}{c}1\\ \mbx\\
\max(0,\overline{\mbz}_{m-1})
\end{array}\right]
\end{array}\right.
\end{array}
\label{DrNet2}
\end{equation}

For notation convenience, we use $\mathsf{DrNet}(m,\Bell;W)$, where $\Bell=[l_1,\cdots,l_m]^T$, to denote the set of functions that can be described by a deep rectifier net of $m$ hidden layers with width $l_i$ for the $i^{th}$ hidden layer and parameter matrix $W$, and we use $\mathsf{DrNet}(m,\Bell)$ to denote the union of all the function nets $\mathsf{DrNet}(m,\Bell;W)$ with a weight matrix $W$ of the structure in (\ref{Wmatrix}).    

Correspondingly, a max-rectifier net, which is associated to a deep rectifier net $\mathsf{DrNet}(m,\Bell;W)$, is defined as 
\begin{equation}
\begin{array}{ll}
({\bf \mathsf{DmrNet}})& 
\left\{\begin{array}{rcl}
f(\mbx) & \triangleq & \dmax_{1\leq j\leq n} f_j(\mbx)\\
f_j(\mbx)&\in&\mathsf{DrNet}(m,\Bell;W)\\
\end{array}\right.
\end{array}
\label{DmrNet}
\end{equation}
and the set of such functions is denoted by $\mathsf{DmrNet}(n,m,\Bell;W)$. When $n=1$, a deep max-rectifier net is reduced to a deep rectifier net. The introduction of a $\max$ operation in the output layer allows one to describe the transformed shallower nets from deeper nets. Next, we will show that the number of linear units in the max-output-layer increases rapidly when a deep rectifier net is transformed to a shallower rectifier net. The following theorem shows that any rectifier network can be transformed to be a shallower network down to the depth of one.

\begin{thm} Let $\Bell=[l_1,l_2,\cdots,l_m]^T\in\mathbb{N}^m$ and $\mathsf{DrNet}(m,\Bell;W)$ be the set of functions that can be represented by a deep rectifier network defined in (\ref{DrNet2}), then, for any $m\geq 2$, the following statements are true:
\begin{enumerate}
\item[i)] Any function that can be represented by a deep rectifier network with $m$ hidden layers can also be realized by 
a max-rectifier net with $(m-1)$ layers, i.e.
\begin{equation}
\begin{array}{c}
\mathsf{DrNet}(m,\Bell)
\subset \mathsf{DmrNet}(2^{l_m},m-1,\hat{\Bell})\\
\end{array} 
\label{Thm1eq1}
\end{equation}
where $\hat{\Bell}=[l_1,l_2,\cdots,l_{m-2},l_{m-1}+l_m]^T\in\mathbb{N}^{m-1}$. 
\item[ii)] Any function that can be represented by a deep rectifier network with $m$ hidden layers can also be realized by a max-rectifier net with a single hidden layer, more precisely
\begin{equation}
\begin{array}{rcl}
\mathsf{DrNet}(m,\Bell)
&\subset& \mathsf{DmrNet}(2^N,1,L) 
\end{array} 
\label{Thm1eq2}
\end{equation}
where $L=\dsum_{i=1}^m l_i, N=\dsum_{i=2}^m (i-1)l_i$. In particular, if all the hidden layer widths $l_i (i\geq 1)$ equal to $l$, then $L=ml$ and $N=\frac{m(m-1)}{2} l=\frac{m-1}{2}L$.
\end{enumerate}
\label{Thm_DrNet}
\end{thm}

The proof of this theorem will be provided in Section 3.3 after we provide three fundamental lemmas in Section 3.1 and Section 3.2.  

\subsection{Two Fundamental Tools }

The following two lemmas (Lemma 1 and Lemma 2) are the basic proposed tools for us to reduce the depth of a deep rectifier neural net. The first shows that the last hidden layer can be removed by adding a max output layer if all the coefficients of linear units are non-negative, while the second lemma is critical to transform the weights of the output layer to be all non-negative. 

\begin{lem}
Let $\mba=[a_1\cdots,a_n]^T\succeq 0$ be an $n$ dimensional vector with non-negative elements, and $\mathbf{f}(\mbx)=[f_1(\mbx),f_2(\mbx),\cdots,f_n(\mbx)]^T$. Then
\begin{equation}
\begin{array}{rcl}
\mba^T\max(0,\mathbf{f})&=&\dmax_{1\leq k\leq N} g_k(\mbx), N=2^n\\
\mathbf{g} &\triangleq& [g_1,g_2,\cdots,g_N]^T\\
&\triangleq& M_n \mathrm{diag}(\mba) \mathbf{f}(\mbx)
\end{array}
\end{equation}
where $\mathrm{diag}(\mathbf{a})=\mathrm{diag} \{a_1,a_2,\cdots,a_n\}$, and $M_n$ is defined recursively as below
\begin{equation}
\begin{array}{rcl}
M_1&=&\left[\begin{array}{c} 0\\ 1\end{array}\right]\\
M_k&=&\left[\begin{array}{cc} M_{k-1}&\mathbf{0}\\ M_{k-1}&\mathbf{1}\end{array}\right], k\geq 2.\\
\end{array}
\end{equation}
\label{fundLem1}
\end{lem} 

{\bf Proof} When $n=1$, $M_1=[0,1]^T$, $\mba=a_1$ is a number and $\mathbf{f}=f_1$ is a 1D function. Therefore $M_1\mathrm{diag}(\mba)\mathbf{f}(\mbx)=[0,a_1]^Tf_1(\mbx)$, that is $g_1=0,g_2=a_1f_1(\mbx)$. Apparently, $a_1\max(0,f_1)=\max(g_1,g_2)$ and {\bf Lemma} \ref{fundLem1} holds when $n=1$. 

Now assume that {\bf Lemma} \ref{fundLem1} holds when $n=k$, by mathematical reduction, we only need to prove that {\bf Lemma} \ref{fundLem1} holds when $n=k+1$. Note that $\mba\succeq 0$ and 
\begin{equation}
\begin{array}{rcl}
\mba^T\max(0,\mathbf{f})&=&\sum_{i=1}^ka_i\max(0,f_i)+ a_{k+1}\max(0,f_{k+1})\\
&=& \max\{h_0,h_0+f_{k+1}\}\\
h_0&\triangleq& \sum_{i=1}^ka_i\max(0,f_i)\\
\end{array}
\label{temLem0}
\end{equation} 
Since {\bf Lemma} \ref{fundLem1} holds when $n=k$, we have
\begin{equation}
h_0=\max\{g_j,j\in [N]\}
\label{temLem1}
\end{equation}
where $N=2^k$ and $[g_1,\cdots,g_N]^T=M_k\mathrm{diag}([a_1,\cdots,a_k]^T) [f_1,\cdots,f_k]^T$, and therefore 
\begin{equation}
\begin{array}{rcl}
h_0+f_{k+1}&=& \max\{g_j+f_{k+1},1\leq j\leq N\}
\end{array}
\label{tempLemma2}
\end{equation} 
Now let $g_{N+j}=g_j+a_{k+1}f_{k+1}$ for $j=1,2,\cdots,N$. Note that 
$M_{k+1}=\left[\begin{array}{cc} M_k&\mathbf{0}\\ M_k&\mathbf{1}\end{array}\right]$, we have 
$[g_1,\cdots,g_{2N}]^T=M_{k+1}\mathrm{diag}([a_1,\cdots,a_{k+1}]^T)\mathbf{f}$.
Then from (\ref{tempLemma2}), (\ref{temLem1}) and (\ref{temLem0}), {\bf Lemma} \ref{fundLem1} holds when $n=k+1$ and the proof is completed.

\begin{flushright}
$\Box$
\end{flushright}

\begin{lem}
Let
\begin{equation}
\begin{array}{rcl}
\mbz_2&=&W_2\max(0,\mbz_1)+\mbb_2\\
\bar{\mbz}_1 &=& W_2\mbz_1+\mbb_2\\
\hat{\mbz}_2&=&-W_2\max(0,\mbz_1)+W_2\mbz_1+\mbb_2\\
\end{array}
\end{equation}
Then we have 
\begin{equation}
\begin{array}{c}
\max(0,\mbz_2)+\max(0,\hat{\mbz}_2)=
\max(0,\mbb_2)+\max\{0,\bar{\mbz}_1)
\end{array}
\label{equiFormula}
\end{equation}
and \begin{equation}
\begin{array}{rcl}
\mba^T\max(0,\mbz_2) &=&\max(0,\mba^T)\max(0,\mbz_2)\\
&&+\max(0,-\mba^T)\max(0,\hat{\mbz}_2)\\
&&-\max(0,-\mba^T)\max(0,\bar{\mbz}_1)\\
&& -\max(0,-\mba^T)\max(0,\mbb_2)\\
\end{array}
\label{non_negative}
\end{equation}
holds for any $\mba\in \mathbb{R}^{l_2}$ where $l_2$ is the dimension of $\mbz_2$.
\label{fundLem2}
\end{lem}

{\bf Proof}. Note that \begin{equation}
\begin{array}{rcl}
\hat{\mbz}_2&=&-W_2\max(0,\mbz_1)+W_2\mbz_1+\mbb_2\\ 
&=& -W_2\max(0,-\mbz_1)+\mbb_2\\
\end{array}
\end{equation}
and 
\begin{equation}
\begin{array}{rcl}
\mbz_2&=&\left\{\begin{array}{ll} W_2\mbz_1+\mbb_2,&\; \mathrm{if}\; \mbz_1\geq 0\\
\mbb_2&\;\mathrm{otherwise}
\end{array}\right.\\
\hat{\mbz}_2&=&\left\{\begin{array}{ll} \mbb_2,&\; \mathrm{if}\; \mbz_1\geq 0\\
W_2\mbz_1+\mbb_2&\;\mathrm{otherwise}
\end{array}\right.
\end{array}
\end{equation}
which implies that, no matter whether $\mbz_1$ is positive or negative, one of $\mbz_2$ and $\hat{\mbz}_2$ is equal to $W_2\mbz_1+\mbb_2$ ($=\bar{\mbz}_1$), and the other is equal to $\mbb_2$. Hence (\ref{equiFormula}) holds and therefore
\begin{equation}
\begin{array}{rcl}
\mba^T\max(0,\mbz_2) &=& \max(0,\mba^T)\max(0,\mbz_2)\\
&&-\max(0,-\mba^T\max(0,\mbz_2)\\
&=&\max(0,\mba^T)\max(0,\mbz_2)\\
&&+\max(0,-\mba^T)\max(0,\hat{\mbz}_2)\\
&&-\max(0,-\mba^T)\max(0,\bar{\mbz}_1)\\
&& -\max(0,-\mba^T)\max(0,\mbb_2)\\
\end{array}
\end{equation}
which proves (\ref{non_negative}) and completes the proof.
\begin{flushright}
$\Box$
\end{flushright}

\subsection{Depth Reduction}

\begin{lem}
Let $f(\mbx)$ be any function in $\mathsf{DrNet}(m,\Bell;W)$ where $m\geq 2$, $\Bell=[l_1,l_2,\cdots,l_m]^T\in\mathbb{N}^m$ and $W$ is defined in (\ref{Wmatrix}). Then there exists $2^{l_m}$ functions, namely $f_j(\mbx)\in \mathsf{DrNet}(m-1,\hat{\Bell};\hat{W})$,  such that 
\begin{equation}
f(\mbx)=\dmax_{1\leq j\leq 2^{l_m}} f_j(\mbx)
\label{claim1Rec}
\end{equation}
where 
\begin{equation}
\begin{small}
\begin{array}{rcl}
\hat{W}&=& \left[\begin{array}{ccccc}\mbb_1& W_{1} &  & & \\ 
\vdots&&\ddots&\\
\mbb_{m-2}&&&W_{m-2}&\\
\hat{\mbb}_{m-1}&&&&\hat{W}_{m-1}\\
\end{array}\right]\\
\end{array}
\end{small}
\label{WmatrixReducedR}
\end{equation}
and
\begin{equation}
\begin{array}{rcl}
\hat{\mbb}_1&=& \left[\begin{array}{c}\mbb_{m-1}\\ \mbb_m+W_m\mbb_{m-1}
\end{array}\right]\\
\hat{W}_{m-1}&=& \left[\begin{array}{c}W_{m-1}\\ W_mW_{m-1}
\end{array}\right].\\
\end{array}
\end{equation}
\label{depthRed1}
\end{lem}

{\bf Proof}: Consider a function $f(\mbx)\in \mathsf{DrNet}(m,\Bell)$, i.e., $f(\mbx)=c+\mba_0^T\mbx+\sum_{k=1}^{m}\mba_k^T\max(0,\mbz_k)$ where $\mbz_k \;$($k\in [m])$ satisfies (\ref{DrNet2}) with $\overline{\mbz}_k$ defined in (\ref{overlineZ}). Denote 
\begin{equation}
f_0(\mbx)=c+\mba_0^T\mbx+\sum_{k=1}^{m-1}\mba_k^T\max(0,\mbz_k).
\end{equation}
Then we have $f(\mbx)=f_0(\mbx)+\mba_m^T\max(0,\mbz_m)$ and $f_0(\mbx)\in \mathrm{DrNet}(m-1,\bar{\Bell},\overline{W})$ where $\bar{\Bell}=[l_1,l_2,\cdots,l_{m-1}]^T$ and 
\begin{equation}
\begin{small}
\begin{array}{rcl}
\overline{W}&=& \left[\begin{array}{ccccc}\mbb_1& W_{1} &  & & \\ 
\vdots&&\ddots&\\
\mbb_{m-2}&&&W_{m-2}&\\
\mbb_{m-1}&&&&W_{m-1}\\
\end{array}\right].\\
\end{array}
\end{small}
\label{Wmatrixm-1}
\end{equation}
Note that $\overline{W}$ is a submatrix of $\hat{W}$ consisting of its first $m-2$ blocks and half of the last block, we have 
\begin{equation}
\mathrm{DrNet}(m-1,\bar{\Bell},\overline{W})\subset \mathrm{DrNet}(m-1,\hat{\Bell},\hat{W})
\end{equation}
and therefore $f_0(\mbx)\in \mathrm{DrNet}(m-1,\hat{\Bell},\hat{W})$. Next we only need to prove that, there exist $2^{l_m}$ functions $g_j(\mbx)\in \mathrm{DrNet}(m-1,\hat{\Bell},\hat{W})$ such that 
\begin{equation}
\begin{array}{rcl}
 \mba_m^T\max(0,\mbz_m) &=&\dmax_{1\leq j\leq 2^{l_m}} g_j(\mbx). 
\end{array}
\label{amZm}
\end{equation}

Let  $\mba_{+}$ denote the subvector of $\mba_m$ consisting of all its positive elements, $\mba_{-}$ the subvector consisting of the remaining non-positive elements, and let $\mbz_{m+},\mbz_{m-}$ be their corresponding subvectors in $\mbz_m$, and $\hat{\mbz}_{m+},\hat{\mbz}_{m-}$ be their corresponding subvectors in $\hat{\mbz}_m$ which is defined as
\begin{equation}
\begin{array}{rcl}
\hat{\mbz}_m&\triangleq&-W_m\max(0,\mbz_{m-1})+W_m\mbz_{m-1}+\mbb_m.\\
\end{array}
\end{equation}

Let $ N=2^{l_m}$ and denote
\begin{equation}
\begin{array}{rcl} 
[\hat{g}_1(\mbx),\cdots,\hat{g}_N(\mbx)]^T &=& M_{l_m} \mathrm{diag}(\mathbf{\tilde{\mba}})\tilde{\mbz}_m\\
\tilde{\mba}&\triangleq&\left[\begin{array}{c}\mba_{+}\\ -\mba_{-}
\end{array}\right]\succeq 0\\
\tilde{\mbz}_m&\triangleq&\left[\begin{array}{c}\mbz_{m+}\\ \hat{\mbz}_{m-}\end{array}\right]\\
\end{array}
\end{equation}
Then from Lemma \ref{fundLem1}, we have
\begin{equation}
\begin{array}{l}
 \max(0,\mba^T)\max(0,\mbz_m)+\max(0,-\mba^T)\max(0,\hat{\mbz}_m)\\
 =\mba_{+}^T\max(0,\mbz_{m+})-\mba_{-}^T\max(0,\mbz_{m-})\\
 =\tilde{\mba}^T\max(0,\tilde{\mbz}_{m})\\
 = \dmax_{1\leq k\leq N} \hat{g}_k(\mbx)\\
 \end{array}
\end{equation}
and by Lemma \ref{fundLem2} it follows   
\begin{equation}
\begin{array}{rcl}
\mba_m^T\max(0,\mbz_m)&=&\max(0,\mba_m^T)\max(0,\mbz_m)\\
&&+\max(0,-\mba_m^T)\max(0,\hat{\mbz}_m)\\
&&-\max(0,-\mba_m^T)\max(0,\tilde{\mbz}_{m-1})\\
&&-\max(0,-\mba_m^T)\max(0,\mbb_m)\\
&=& \dmax_{1\leq k\leq N} \hat{g}_k(\mbx)\\
&&-\max(0,-\mba_m^T)\max(0,\tilde{\mbz}_{m-1})\\
&&-\max(0,-\mba_m^T)\max(0,\mbb_m)\\
\end{array}
\label{amZm2}
\end{equation}
where 
\begin{equation}
\begin{array}{rcl}
\tilde{\mbz}_{m-1}&\triangleq& W_m\mbz_{m-1}+\mbb_m.
\end{array}
\end{equation}

Note that $-\max(0,-\mba_m^T)\max(0,\mbb_m)$ is a scalar number, and each of the elements of $\tilde{\mbz}_m$ and $\tilde{\mbz}_{m-1}$ is a function in $\mathrm{DrNet}(m-1,\hat{\Bell},\hat{W})$, we have 

\begin{equation}
\begin{array}{rcl}
g_j(\mbx) &\triangleq& \hat{g}_j(\mbx)-\max(0,-\mba_m^T)\max(0,\tilde{\mbz}_{m-1})\\
&&-\max(0,-\mba_m^T)\max(0,\mbb_m)\\
&\in& \mathrm{DrNet}(m-1,\hat{\Bell},\hat{W}). 
\end{array}
\end{equation} 

Thus, (\ref{amZm}) follows from (\ref{amZm2}), and the proof is completed. 

\begin{flushright}
$\Box$
\end{flushright}

\subsection{Proof of Theorem 1}

The first statement (\ref{Thm1eq1}) follows directly from {\bf Lemma} \ref{depthRed1}. Next we prove (\ref{Thm1eq2}) by using mathematical reduction method. When $m=2$, we have $L=l_2+l_1, N=l_2$, (\ref{Thm1eq2}) is identical to (\ref{Thm1eq1}) and thus holds when  $m=2$.

Now assume that (\ref{Thm1eq2}) holds when $m=k$ for some $k\geq 2$. For $m=k+1$, (\ref{Thm1eq1}) implies that 
\begin{equation}
\begin{array}{c}
\mathsf{DrNet}(k+1,\Bell )\subset \mathsf{DmrNet}(2^{l_{k+1}},k,\hat{\Bell})\\
\end{array} 
\label{tempThm1}
\end{equation}
where 
\begin{equation}
\begin{array}{rcl}
\Bell &=& [l_1,l_2,\cdots,l_{k+1}]^T\\
\hat{\Bell}&=& [l_1,l_2,\cdots,l_{k-1},l_{k}+l_{k+1}]^T.
\end{array}
\end{equation}
Since (\ref{Thm1eq2}) holds when $m=k$, we have 
\begin{equation}
\begin{array}{c}
\mathsf{DrNet}(k,\hat{\Bell}) \subset \mathsf{DmrNet}(2^{\hat{N}},1,L)
\end{array} 
\label{tempThm2}
\end{equation}
where $L=\dsum_{i=1}^{k+1} l_i$ and $\hat{N} = (k-1)(l_k+l_{k+1})+\sum_{i=2}^{k-1} (i-1)l_i = -l_{k+1}+\sum_{i=2}^{k+1} (i-1)l_i$. 

Let $f(\mbx)$ be a function in $\mathsf{DrNet}(k+1,\Bell)$. From (\ref{tempThm1}), there exist $2^{l_{k+1}}$ functions, namely $f_j(\mbx)$, in $\mathsf{DrNet}(k,\hat{\Bell})$, such that 
\begin{equation}
f(\mbx)=\dmax_{1\leq j\leq 2^{l_{k+1}}} f_j(\mbx).
\end{equation}

For each $f_j(\mbx)\in \mathsf{DrNet}(k,\hat{\Bell})$, from (\ref{tempThm2}), there exist $2^{\hat{N}}$ functions, namely $g_{ji}(\mbx)\in \mathsf{DrNet}(1,L)$ such that 
\begin{equation}
f_j(\mbx)=\dmax_{1\leq i\leq 2^{\hat{N}}} g_{ji}(\mbx)
\end{equation}
and therefore
\begin{equation}
\begin{array}{rcl}
f(\mbx)&=&\dmax_{1\leq j\leq 2^{l_{k+1}}} \dmax_{1\leq i\leq 2^{\hat{N}}} g_{ji}(\mbx)\\
&\in& \mathsf{DmrNet}(2^N,1,L)
\end{array}
\end{equation}
where 
\begin{equation}
N=\hat{N}+l_{k+1}=\sum_{i=2}^{k+1} (i-1)l_i.
\end{equation}

Note that $f(\mbx)$ can be any function in  $\mathsf{DrNet}(k+1,\Bell)$, so $\mathsf{DrNet}(k+1,\Bell)\subset \mathsf{DmrNet}(2^N,1,L)$ which proves (\ref{Thm1eq2}) for $m=k+1$ and completes the proof.
\begin{flushright}
$\Box$
\end{flushright}

 \section{Transformations of Deep Rectifier Nets with Full Skip Connections}

Consider the following model of a general rectifier net with depth $m$  as below
\begin{equation}
\begin{array}{ll}
({\bf \mathrm{DrNet}})& 
\left\{\begin{array}{rcl}
f(\mbx) &=& b_0+\mba_0^T\mbx + \dsum_{i=1}^m \mba_i^T\max(0,\mbz_i)\\
\mbz_1&=&W_{1,0}\mbx+\mbb_1\\
\mbz_k&=&\dsum_{i=1}^{k-1} W_{k,i} \max(0,\mbz_i)\\
&& +W_{k,0}\mbx+\mbb_k, k=2,\cdots,m.\\
\end{array}\right.
\end{array}
\label{DrNetGeneral}
\end{equation}

In this model, the output of the $i^{th}$ layer is $\max(0,\mbz_i)\in\mathbb{R}^{l_i}$, where $l_i$ denote the number of units of the $i^{th}$ layer. The input $\mbx$ is treated as the output of the $0^{th}$ layer and its dimension is denoted by $l_0$. Correspondingly, $W_{i,j}\in\mathbb{R}^{l_i\times l_j}$ and $\mbb_i\in\mathbb{R}^{l_i}$ denotes the weight matrix and the bias vector of the linear units from the $j^{th}$ layer to the $i^{th}$ layer. In traditional rectifier networks, only the adjacent layers are connected, that is, $W_{i,j}=0$ for any $i>j+1$, and $W$ is then a block diagonal matrix.

The final output of the network is the maximum of several linear functions of the input and the hidden layer outputs. The outputs of hidden layers can be viewed as generated nonlinear features for nonlinear function representation. In traditional rectifier network, the final output is usually a linear function of the output of the last hidden layer. In this case, $n=1$ and $\mba_{1,i}=0$ for any $i<m$.  

Let
\begin{equation}
\begin{array}{rcl}
W&=& \left[\begin{array}{ccccc}\mbb_1& W_{1,0} &  & & \\ \mbb_2& W_{2,0}& W_{2,1}  & &\\
\vdots&\vdots&\vdots&\ddots&\\
\mbb_m&W_{m,0}&W_{m,1}&\cdots&W_{m,m-1}\\
\end{array}\right]
\end{array}
\label{WmatrixGeneral}
\end{equation}
where empty blocks are zero blocks with proper dimensions,
and denote
\begin{equation}
\begin{array}{rcl}
\overline{\mbz}_k &=& \left[\begin{array}{c} \mbz_1\\ \mbz_2 \\ \vdots\\\mbz_k
\end{array}\right],
\end{array} 
\begin{array}{rcl}
\overline{\mba}_k &=& \left[\begin{array}{c}\mba_1 \\ \mba_2\\ \vdots\\ \mba_k
\end{array}\right]
\end{array} 
\end{equation}
for $k=1,2,\cdots,m$. Then the formulation (\ref{DrNetGeneral}) can be simplified as 
\begin{equation}
\begin{array}{ll}
({\bf \mathrm{DrNet}})& 
\left\{\begin{array}{rcl}
f(\mbx) &=& b_0+\mba_0^T\mbx + \overline{\mba}_m^T\max(0,\overline{\mbz}_m)\\
\overline{\mbz}_m&=&W\left[\begin{array}{c}1\\ \mbx\\ \max(0,\overline{\mbz}_{m-1})
\end{array}\right]
\end{array}\right.
\end{array}
\label{DrNetGeneral2}
\end{equation}

The set of functions that can be described by a deep rectifier net with $m$ hidden layers of width $l_i$ for the $i^{th}$ hidden layer, and a weight matrix $W$, is denoted by $\mathsf{DrNet}(m,\Bell;W)$ where $\Bell=[l_1,l_2,\cdots,l_m]^T$ is the vector of the hidden layer widths.   

\begin{thm} Let $\mathsf{DrNet}(m,\Bell;W)$ be the set of functions that can be represented by a deep rectifier network defined in (\ref{DrNet}) and $m\geq 2$, then the following statements are true:
\begin{enumerate}
\item[i)] For any function $f(\mbx)\in \mathsf{DrNet}(m,\Bell;W)$, there exists $2^{l_m}$ functions, namely $f_j(\mbx)\in \mathsf{DrNet}(m-1,\hat{\Bell};\hat{W})$,  such that 
\begin{equation}
f(\mbx)=\dmax_{1\leq j\leq 2^{l_m}} f_j(\mbx)
\label{claim1}
\end{equation}
where $\hat{\Bell}=[l_1,\cdots, l_{m-2}, l_{m-1}+2l_m]^T$ and
\begin{equation}
\begin{small}
\begin{array}{rcl}
\hat{W}&=& \left[\begin{array}{cccc}\mbb_1& W_{1,0}   & & \\
\vdots&\vdots&\ddots&\\
\mbb_{m-1}&W_{m-1,0}&\cdots&W_{m-1,m-2}\\
\mbb_m&W_{m,0}&\cdots&W_{m,m-2}\\
\bar{\mbb}_m&\bar{W}_{m,0}&\cdots&\bar{W}_{m,m-2}\\
\end{array}\right]\\
\bar{\mbb}_m&=&\mbb_m+W_{m,m-1}\mbb_{m-1}\\
\bar{W}_{m,k}&=& W_{m,k}+W_{m,m-1}W_{m-1,k} 
\end{array}
\end{small}
\label{WmatrixReducedGeneral}
\end{equation}

Denote 
\begin{equation}
\begin{array}{c}
\mathsf{DmrNet}(n,m,\Bell;W)\\
\triangleq \left\{\max_{j\in[n]} f_j(\mbx):f_j(\mbx)\in \mathsf{DrNet}(n,m,\Bell);W\right\}.
\end{array}
\label{defDrmNet}
\end{equation}
Then we have  
\begin{equation}
\begin{array}{c}
\mathsf{DrNet}(m,\Bell;W)\subset \mathsf{DrmNet}(2^{l_m},m-1,\hat{\Bell};\hat{W}).
\end{array} 
\label{depthReductionGeneral}
\end{equation}
\item[ii)] Any function that can be represented by a deep rectifier network with $m$ hidden layers can also be realized by a max-rectifier network with only one hidden layer, more precisely, there exist $\tilde{W}=[\tilde{\mbb}_1,\tilde{W}_{0,1}]\in \mathbb{R}^{L\times (l_0+1)}$,  where $l_0$ is the dimension of $\mbx$ and $L=\dsum_{i=1}^m 2^{i-1}l_i$ such that
\begin{equation}
\begin{array}{rcl}
\mathsf{DrNet}(m,\Bell;W)&\subset& \mathsf{DmrNet}(2^N,1,L;\tilde{W})
\end{array} 
\label{Dr2Br}
\end{equation}
where $N = \dsum_{i=2}^m (2^{i-1}-1)l_i$. In particular, if all the hidden layer widths $l_i (i\geq 1)$ equal to $l$, then $L=(2^m-1)l$ and $N=(2^m-m-1)l=L-ml$. 
\end{enumerate}
\end{thm}

Theorem 5 shows that any function represented by a deep rectifier network can also be realized by a shallow net with only one hidden layer. However, with the same number number of units, a deep rectifier net is exponentially more efficient in creating the number of maxout units in the output layer and exponentially more efficient in creating the number of nonlinear features than the corresponding shallow net.

{\bf Proof}: Let 
\begin{equation}
f(\mbx)=f_0(\mbx)+\mba_{m-1}^T\max(0,\mbz_{m-1})+\mba_m^T\max(0,\mbz_m)
\end{equation}
 be a function in $\mathsf{DrNet}(1,m,\Bell;W)$ where 
\begin{equation}
f_0(\mbx)=b_0+\mba_0^T\mbx+\sum_{i=1}^{m-2}\mba_i^T\max(0,\mbz_i)
\end{equation} 
Next, we will show how to remove the term $\max(0,\mbz_m)$ and add some new nodes in the $(m-1)^{th}$ layer so that $f(\mbx)$ can be represented by a rectifier network with $(m-1)$ layers. Let 
\begin{equation}
\begin{array}{rcl}
\mbz_m^0 &\triangleq& \mbz_m - W_{m,m-1}\max(0,\mbz_{m-1})\\
\hat{\mbz}_m &\triangleq& \mbz_m^0-W_{m,m-1}\max(0,-\mbz_{m-1})\\
\end{array}
\end{equation}
Note that $\mbz_m=\mbz_m^0 + W_{m,m-1}\max(0,\mbz_{m-1})$ and 
\begin{equation}
\begin{array}{rcl}
\mbz_m&=&\left\{\begin{array}{ll} \mbz_{m}^0+W_{m,m-1}\mbz_{m-1},&\; \mathrm{if}\; \mbz_{m-1}\geq 0\\
\mbz_{m}^0&\;\mathrm{otherwise}
\end{array}\right.\\
\hat{\mbz}_m&=&\left\{\begin{array}{ll} \mbz_{m}^0,&\; \mathrm{if}\; \mbz_1\geq 0\\
\mbz_{m}^0+W_{m,m-1}\mbz_{m-1}&\;\mathrm{otherwise}
\end{array}\right.
\end{array}
\end{equation}
which imply that, no matter whether $\mbz_{m-1}$ is positive or negative, one of $\mbz_m$ and $\hat{\mbz}_m$ is equal to $\mbz_{m}^0+W_{m,m-1}\mbz_{m-1}$, and the other is equal to $\mbz_{m}^0$. Hence
\begin{equation}
\begin{array}{c}
\max(0,\mbz_m)+\max(0,\hat{\mbz}_m)=\\ \max(0,\mbz_m^0)+
 \max(0,\mbz_m^0+W_{m,m-1}\mbz_{m-1})
\end{array}
\end{equation}
and therefore
\begin{equation}
\begin{array}{rcl}
f(\mbx)&=&f_0(\mbx)+\mba_{m-1}^T\max(0,\mbz_{m-1})\\
&&+\{\max(0,\mba_m^T)-\max(0,-\mba_m^T)\}\max(0,\mbz_m)\\
&=&f_0(\mbx)+\mba_{m-1}^T\max(0,\mbz_{m-1})\\
&&+\max(0,\mba_m^T)\max(0,\mbz_m)\\
&&+\max(0,-\mba_m^T)\max(0,\hat{\mbz}_m)\\
&&-\max(0,\mba_m^T)\max(0,\mbz_m^0)\\
&&-\max(0,\mba_m^T)\max(0,\mbz_m^0+W_{m,m-1}\mbz_{m-1})\\
&=& f_0(\mbx) + \hat{\mba}_{m-1}^T\max(0,\hat{\mbz}_{m-1})\\
&&  + [\max(0,\mba_m^T),\max(0,-\mba_m^T)]\left[\begin{array}{c}
\max(0,\mbz_m)\\ \max(0,\hat{\mbz}_m)
\end{array}\right]\\
&=& f_0(\mbx) + \hat{\mba}_{m-1}^T\max(0,\hat{\mbz}_{m-1})\\
&&  + \mathbf{1}_m^T P\left[\begin{array}{c}
\max(0,\mbz_m)\\ \max(0,\hat{\mbz}_m)
\end{array}\right]\\
\end{array}
\end{equation}
where $\mathbf{1}_m$ is a vector with all elements being 1, and
\begin{equation}
\begin{array}{rcl}
P&\triangleq& \left[\mathrm{diag}\{\max(0,\mba_m)\}, \mathrm{diag}\{\max(0,-\mba_m)\}\right]\\
\hat{\mba}_{m-1}^T &\triangleq& [\mba_{m-1}^T,-\max(0,-\mba_m^T),-\max(0,-\mba_m^T)]\\
\hat{\mbz}_{m-1} &\triangleq& \left[\begin{array}{c}
\mbz_{m-1}\\ \mbz_m^0\\ \mbz_m^0+W_{m,m-1}\mbz_{m-1}\\
\end{array}\right]\\
&=&\left[\begin{array}{cccc}
\mbb_{m-1}& W_{m-1,0}& \cdots &W_{m-1,m-2}\\
\mbb_{m}& W_{m,0}& \cdots &W_{m,m-2}\\
\hat{\mbb}_{m}& \hat{W}_{m,0}& \cdots &\hat{W}_{m,m-2}\\
\end{array}\right]\overline{\mbz}_{m-2}\\
\hat{\mbb}_m &\triangleq& \mbb_m+W_{m,m-1}\mbb_{m-1}\\
\hat{W}_{m,i}&\triangleq&W_{m,i}+W_{m,m-1}W_{m-1,i}.
\end{array}
\label{hatZm-1}
\end{equation}

Let $\hat{\mbz}_i=z_i$ for $i=1,2,\cdots,m-2$, $\hat{\mbz}_{m-1}$ be defined as in  (\ref{hatZm-1}), and
\begin{equation}
\overline{\hat{\mbz}}_k \triangleq \left[\begin{array}{c}\hat{\mbz}_1\\
\hat{\mbz}_2\\ \vdots\\ \hat{\mbz}_k\\
\end{array}\right].
\end{equation}
Then we have \begin{equation}
\overline{\hat{\mbz}}_{m-1}=\hat{W} \left[ \begin{array}{c}1\\ \mbx\\
\max(0,\overline{\hat{\mbz}}_{m-2})
\end{array}\right]
\end{equation}
and therefore
\begin{equation}
f_0(\mbx) + \hat{\mba}_{m-1}^T\max(0,\hat{\mbz}_{m-1})\in \mathsf{DrNet}(m-1,\hat{l},\hat{W}). 
\end{equation}

By Lemma \ref{fundLem1}, 
\begin{equation}
\begin{array}{c}
\mathbf{1}_m^T P\left[\begin{array}{c}
\max(0,\mbz_m)\\ \max(0,\hat{\mbz}_m)
\end{array}\right]=\dmax_{1\leq j\leq N} g_j(\mbx)
\end{array}
\end{equation}
where $N=2^m$ and 
\begin{equation}
[g_1,g_2,\cdots, g_N]=M_mP\left[\begin{array}{c}
\mbz_m\\ \hat{\mbz}_m\end{array}\right].
\end{equation}

Note that 
\begin{equation}
\begin{array}{rcl}
\left[\begin{array}{c}\mbz_m\\ \hat{\mbz}_m
\end{array}\right]&=&\left[\begin{array}{cccc}
\mbb_m&W_{m,0}&\cdots&W_{m,m-2}\\
\hat{\mbb}_m & \hat{W}_{m,0}&\cdots&\hat{W}_{m,m-2}
\end{array}\right]\overline{\mbz}_{m-2}\\
&&+\left[\begin{array}{c}
 W_{m,m-1}\\ -W_{m,m-1}
\end{array}\right]\max(0,\mbz_{m-1})\\
\end{array}
\end{equation}
That is, each element of $\mbz_m$ and $\hat{\mbz}_m$ is a linear function on 
\begin{equation}
 \left[ \begin{array}{c}1\\ \mbx\\
\max(0,\overline{\mbz}_{m-1})
\end{array}\right]
\end{equation}
and therefore a linear function on 
\begin{equation}
\left[ \begin{array}{c}1\\ \mbx\\
\max(0,\overline{\hat{\mbz}}_{m-1})
\end{array}\right]
\end{equation}
since
\begin{equation}
\overline{\hat{\mbz}}_{m-1}=\left[\begin{array}{c}\overline{\mbz}_{m-1}\\ \mbz_m^0\\ \mbz_m^0+W_{m,m-1}\mbz_{m-1}
\end{array}\right].
\end{equation}
Hence $f_j(\mbx)\triangleq f_0(\mbx)+g_j(\mbx)\in \mathsf{DrNet}(m-1,\hat{l},\hat{W})$, and (\ref{claim1}) holds. (\ref{depthReductionGeneral}) follows straightforward from (\ref{claim1}) and the definition of $\mathsf{DmrNet}$ in (\ref{defDrmNet}).

Next we prove (\ref{Dr2Br}) by using mathematical reduction method.

When $m=2$, we have $L=2l_2+l_1, N=l_2$, (\ref{Dr2Br}) is identical to (\ref{depthReductionGeneral}) and thus (\ref{Dr2Br}) holds if $m=2$.

Now assume that (\ref{Dr2Br}) is true when $m=k$ for some $k\geq 2$, 
we will prove that (\ref{Dr2Br}) is true when $m=k+1$.   
Apply (\ref{depthReductionGeneral}) for the case $m=k+1$, we have 
\begin{equation}
\begin{array}{c}
\mathsf{DrNet}(k+1,l_1,l_2,\cdots,l_{k+1})\subset\\ \mathsf{DmrNet}(2^{l_{k+1}},k,l_1,l_2,\cdots,l_{k-1},l_{k}+2l_{k+1})\\
\end{array} 
\label{temp1}
\end{equation}
Now apply (\ref{Dr2Br}) on the case $m=k$, we have 
\begin{equation}
\begin{array}{r}
\mathsf{DmrNet}(2^{l_{k+1}},k,l_1,l_2,\cdots,l_{k-1},l_{k}+2l_{k+1})\\\subset \mathsf{DmrNet}(2^N,1,L)
\end{array} 
\label{temp2}
\end{equation}
where $L=\dsum_{i=1}^{k-1} 2^{i-1}l_i+2^{k-1}(l_k+2l_{k+1})=\dsum_{i=1}^{k+1} 2^{i-1}l_i$ and $N = l_{k+1} + \dsum_{i=2}^{k-1} (2^{i-1}-1)l_i+(2^{k-1}-1)(l_k+2l_{k+1})=\dsum_{i=2}^{k+1} (2^{i-1}-1)l_i$. Then from (\ref{temp1}) and (\ref{temp2}), it follows that (\ref{Dr2Br}) is true for $m=k+1$ as well. Hence, by mathematical reduction, the proof of (\ref{Dr2Br}) is completed.
\begin{flushright}
$\Box$
\end{flushright}

\section{Transformations of Deep Residual Networks}

In deep residual networks, skip connections are added upon the traditional convolution neural nets. In a typical residual net, two adjacent layers are grouped together to formulate a residual unit, and a skip connection is added between two adjacent residual units with an identity or projective map to bypass one residual unit \cite{he2016identity}. 

Consider the following model of a deep residual rectifier net with depth $m$  as below
\begin{equation}
\begin{array}{ll}
({\bf \mathrm{DresNet}})& 
\left\{\begin{array}{rcl}
f(\mbx) & \triangleq &c+\mba_0^T\mbx+\dsum_{k=1}^m \mba_k^T\max(0,\mbz_k)\\
\mbz_1 &=& W_1\mbx+\mbb_1\\
\mbz_{2k} &=& W_{2k}\max(0,\mbz_{2k})+\mbb_{2k}\\
\mbz_{2k+1} &=& W_{2k+1}\max(0,\mbz_{2k})\\
&&+A_{2k+1}\max(0,\mbz_{2k-1})+\mbb_{2k+1}\\
\end{array}\right.
\end{array}
\label{DresNet}
\end{equation}
which can also be modelled as 
\begin{equation}
\begin{array}{ll}
({\bf \mathrm{DresNet}})& 
\left\{\begin{array}{rcl}
f(\mbx) & \triangleq &c+\mba_0^T\mbx+\dsum_{k=1}^m \mba_k^T\max(0,\mbz_k)\\
\overline{\mbz}_m &=& W\left[\begin{array}{c}1\\ \mbx\\ \max(0,\overline{\mbz}_{m-1}
\end{array}\right]\\
\end{array}\right.
\end{array}
\label{DresNet2}
\end{equation}
where $\overline{\mbz}_k$ is defined in (\ref{overlineZ}), and $W$ is defined as below 
\begin{equation}
\begin{array}{l}
\left[\begin{array}{cccccc}\mbb_1& W_{1} &  & & &\\ \mbb_2& A_2 & W_{2}  & &&\\
\mbb_3&&A_3&W_3\\
\vdots&&&\ddots&\ddots\\
\mbb_m&&&&A_m&W_m\\
\end{array}\right]\\
A_k=\begin{array}{ll} 0,&\mathrm{if}\; k \; \mathrm{is\; even}\\
 \end{array}.
\end{array}
\end{equation}

 Next we consider the transformations of residual nets and show their superior expressive power compared to the plain rectifier networks (with no skip connections).
 
\begin{thm} Let $\mathsf{DresNet}(m,\Bell;W)$ be the set of all the functions that can be represented by a deep residual network defined in (\ref{DresNet2}) and $m\geq 2$, then for any function $f(\mbx)\in \mathsf{DresNet}(m,\Bell;W)$, there exists $2^{l_m}$ functions, namely $f_j(\mbx)\in \mathsf{DresNet}(m-1,\hat{\Bell};\hat{W})$,  such that 
\begin{equation}
f(\mbx)=\dmax_{1\leq j\leq 2^{l_m}} f_j(\mbx)
\label{claim1}
\end{equation}
where  $\hat{W}$ is defined as
\begin{equation}
\left[\begin{array}{cccccc}\mbb_1& W_{1} &  & & &\\ \mbb_2& A_2 & W_{2}  & &&\\
\vdots&&\ddots&\ddots&\\
\mbb_{m-2}&&&A_{m-2}&W_{m-2}&\\
\hat{\mbb}_{m-1}&&&&\hat{A}_{m-1}&\hat{W}_{m-1}\\
\end{array}\right]
\end{equation}
and $\hat{\Bell}$, $\hat{W}_{m-1}$, $\hat{A}_{m-1}$ and $\hat{\mbb}_{m-1}$ are defined below separately for the cases when $m$ is even or odd:
\begin{enumerate}
\item[i)] When $m$ is even,     
\begin{equation}
\begin{array}{rcl}
\hat{\Bell} &=& [l_1,\cdots, l_{m-2},l_{m-1}+\l_m]^T\\
\hat{\mbb}_{m-1}&=&\left[\begin{array}{c}\mbb_{m-1}\\\mbb_m+W_m\mbb_{m-1}
\end{array}\right]\\
\hat{W}_{m-1} &=& \left[\begin{array}{c} W_{m-1}\\ W_mW_{m-1}\end{array}\right]\\
\hat{A}_{m-1} &=& \left[\begin{array}{c} A_{m-1}\\ W_mA_{m-1}\end{array}\right]\\
\end{array}
\label{WmatrixReducedResidualEven}
\end{equation}

\item[ii)] When $m$ is odd, 
\begin{equation}
\begin{array}{rcl}
\hat{\Bell} &=& [l_1,\cdots, l_{m-2},l_{m-1}+2\l_m]^T\\
\hat{\mbb}_{m-1}&=&\left[\begin{array}{c}\mbb_{m-1}\\ \mbb_m\\ \mbb_m+W_m\mbb_{m-1}
\end{array}\right]\\
\hat{W}_{m-1} &=& \left[\begin{array}{c} W_{m-1}\\ A_m\\  W_mW_{m-1}\end{array}\right]\\
\hat{A}_{m-1} &=& \mathbf{0}.\\
\end{array}
\label{WmatrixReducedResidualOdd}
\end{equation}
\end{enumerate}
\end{thm}

{\bf Proof}: Let 
\begin{equation}
f(\mbx)=f_0(\mbx)+\mba_{m-1}^T\max(0,\mbz_{m-1})+\mba_m^T\max(0,\mbz_m)
\end{equation}
 be a function in $\mathsf{DresNet}(m,\Bell;W)$ where 
\begin{equation}
f_0(\mbx)=b_0+\mba_0^T\mbx+\sum_{i=1}^{m-2}\mba_i^T\max(0,\mbz_i)
\end{equation} 
Next, we will show how to remove the term $\max(0,\mbz_m)$ and add some new nodes in the $(m-1)^{th}$ layer so that $f(\mbx)$ can be represented by a rectifier network with $(m-1)$ layers. Let 
\begin{equation}
\begin{array}{rcl}
\mbz_m^0 &\triangleq& \mbz_m - W_{m}\max(0,\mbz_{m-1})\\
\hat{\mbz}_m &\triangleq& \mbz_m^0-W_{m}\max(0,-\mbz_{m-1})\\
\end{array}
\end{equation}
Note that $\mbz_m=\mbz_m^0 + W_{m}\max(0,\mbz_{m-1})$ and 
\begin{equation}
\begin{array}{rcl}
\mbz_m&=&\left\{\begin{array}{ll} \mbz_{m}^0+W_{m}\mbz_{m-1},&\; \mathrm{if}\; \mbz_{m-1}\geq 0\\
\mbz_{m}^0&\;\mathrm{otherwise}
\end{array}\right.\\
\hat{\mbz}_m&=&\left\{\begin{array}{ll} \mbz_{m}^0,&\; \mathrm{if}\; \mbz_1\geq 0\\
\mbz_{m}^0+W_{m}\mbz_{m-1}&\;\mathrm{otherwise}
\end{array}\right.
\end{array}
\end{equation}
which imply that, no matter whether $\mbz_{m-1}$ is positive or negative, one of $\mbz_m$ and $\hat{\mbz}_m$ is equal to $\mbz_{m}^0+W_{m}\mbz_{m-1}$, and the other is equal to $\mbz_{m}^0$. Hence
\begin{equation}
\begin{array}{c}
\max(0,\mbz_m)+\max(0,\hat{\mbz}_m)=\\ \max(0,\mbz_m^0)+
 \max(0,\mbz_m^0+W_{m}\mbz_{m-1})
\end{array}
\end{equation}
and therefore
\begin{equation}
\begin{array}{rcl}
f(\mbx)&=&f_0(\mbx)+\mba_{m-1}^T\max(0,\mbz_{m-1})\\
&&+\{\max(0,\mba_m^T)-\max(0,-\mba_m^T)\}\max(0,\mbz_m)\\
&=&f_0(\mbx)+\mba_{m-1}^T\max(0,\mbz_{m-1})\\
&&+\max(0,\mba_m^T)\max(0,\mbz_m)\\
&&+\max(0,-\mba_m^T)\max(0,\hat{\mbz}_m)\\
&&-\max(0,\mba_m^T)\max(0,\mbz_m^0)\\
&&-\max(0,\mba_m^T)\max(0,\mbz_m^0+W_{m}\mbz_{m-1})\\
&=& f_0(\mbx) + \hat{\mba}_{m-1}^T\max(0,\hat{\mbz}_{m-1})\\
&&  + [\max(0,\mba_m^T),\max(0,-\mba_m^T)]\left[\begin{array}{c}
\max(0,\mbz_m)\\ \max(0,\hat{\mbz}_m)
\end{array}\right]\\
&=& f_0(\mbx) + \hat{\mba}_{m-1}^T\max(0,\hat{\mbz}_{m-1})\\
&&  + \mathbf{1}_m^T P\left[\begin{array}{c}
\max(0,\mbz_m)\\ \max(0,\hat{\mbz}_m)
\end{array}\right]\\
\end{array}
\end{equation}
where $\mathbf{1}_m$ is a vector with all elements being 1, and
\begin{equation}
\begin{array}{rcl}
P&\triangleq& \left[\mathrm{diag}\{\max(0,\mba_m)\}, \mathrm{diag}\{\max(0,-\mba_m)\}\right]\\
\hat{\mba}_{m-1}^T &\triangleq& [\mba_{m-1}^T,-\max(0,-\mba_m^T),-\max(0,-\mba_m^T)]\\
\hat{\mbz}_{m-1} &\triangleq& \left[\begin{array}{c}
\mbz_{m-1}\\ \mbz_m^0\\ \mbz_m^0+W_{m}\mbz_{m-1}\\
\end{array}\right]\\
&=&\left[\begin{array}{ccccc}
\mbb_{m-1}& \cdots &A_{m-1}&W_{m-1}\\
\mbb_{m}& \cdots &\mathbf{0}&A_m\\
\hat{\mbb}_{m}& \cdots &\hat{A}_m&\hat{W}_{m}\\
\end{array}\right]\overline{\mbz}_{m-2}\\
\hat{\mbb}_m &\triangleq& \mbb_m+W_{m}\mbb_{m-1}\\
\hat{A}_{m}&\triangleq&W_{m}A_{m-1}\\
\hat{W}_{m}&\triangleq&W_{m}W_{m-1}+A_m\\
\end{array}
\label{hatZm-1Res}
\end{equation}

Let $\hat{\mbz}_i=z_i$ for $i=1,2,\cdots,m-2$, $\hat{\mbz}_{m-1}$ be defined as in  (\ref{hatZm-1Res}), and
\begin{equation}
\overline{\hat{\mbz}}_k \triangleq \left[\begin{array}{c}\hat{\mbz}_1\\
\hat{\mbz}_2\\ \vdots\\ \hat{\mbz}_k\\
\end{array}\right].
\end{equation}
Then we have \begin{equation}
\overline{\hat{\mbz}}_{m-1}=\hat{W} \left[ \begin{array}{c}1\\ \mbx\\
\max(0,\overline{\hat{\mbz}}_{m-2})
\end{array}\right]
\end{equation}
and therefore
\begin{equation}
f_0(\mbx) + \hat{\mba}_{m-1}^T\max(0,\hat{\mbz}_{m-1})\in \mathsf{DrNet}(m-1,\hat{l},\hat{W}). 
\end{equation}

By Lemma \ref{fundLem1}, 
\begin{equation}
\begin{array}{c}
\mathbf{1}_m^T P\left[\begin{array}{c}
\max(0,\mbz_m)\\ \max(0,\hat{\mbz}_m)
\end{array}\right]=\dmax_{1\leq j\leq N} g_j(\mbx)
\end{array}
\end{equation}
where $N=2^m$ and 
\begin{equation}
[g_1,g_2,\cdots, g_N]=M_mP\left[\begin{array}{c}
\mbz_m\\ \hat{\mbz}_m\end{array}\right].
\end{equation}

Note that 
\begin{equation}
\begin{array}{rcl}
\left[\begin{array}{c}\mbz_m\\ \hat{\mbz}_m
\end{array}\right]&=&\left[\begin{array}{cccc}
\mbb_m&\cdots&&A_{m-1}\\
\hat{\mbb}_m &\cdots&&\hat{A}_{m}
\end{array}\right]\overline{\mbz}_{m-2}\\
&&+\left[\begin{array}{c}
 W_{m}\\ -W_{m}
\end{array}\right]\max(0,\mbz_{m-1})\\
\end{array}
\end{equation}
which shows that each element of $\mbz_m$ and $\hat{\mbz}_m$ is a linear function of 
\begin{equation}
 \left[ \begin{array}{c}1\\ \mbx\\
\max(0,\overline{\mbz}_{m-1})
\end{array}\right]
\end{equation}
and therefore a linear function of
\begin{equation}
\left[ \begin{array}{c}1\\ \mbx\\
\max(0,\overline{\hat{\mbz}}_{m-1})
\end{array}\right]
\end{equation}
since
\begin{equation}
\overline{\hat{\mbz}}_{m-1}=\left[\begin{array}{c}\overline{\mbz}_{m-1}\\ \mbz_m^0\\ \mbz_m^0+W_{m}\mbz_{m-1}
\end{array}\right].
\end{equation}

Note that when $m$ is odd, we have $A_{m-1}=0, \hat{A}_m=W_mA_{m-1}=0$ and 
hence $f_j(\mbx)\triangleq f_0(\mbx)+g_j(\mbx)\in \mathsf{DresNet}(m-1,\hat{\Bell},\hat{W})$ where $\hat{\Bell}$ and $\hat{W}$ are defined in (\ref{WmatrixReducedResidualOdd}). Similarly, when $m$ is even, we have $A_m=0$ and the middle block of $\hat{\mbz}_{m-1}$ in (\ref{hatZm-1Res}) has only the bias vector and therefore the hidden nodes associated with this block are empty. Hence $f_j(\mbx)\triangleq f_0(\mbx)+g_j(\mbx)\in \mathsf{DresNet}(m-1,\hat{\Bell},\hat{W})$ but $\hat{\Bell}$ and $\hat{W}$ are defined in (\ref{WmatrixReducedResidualEven}).  
\begin{flushright}
$\Box$
\end{flushright}

\begin{thm}
Denote 
\begin{equation}
\begin{array}{c}
\mathsf{DmresNet}(n,m,\Bell;W)\\
\triangleq \left\{f(\mbx)=\dmax_{1\leq n\leq n} f_j(\mbx):f_j(\mbx)\in \mathsf{DresNet}(n,m,\Bell;W\right\}.
\end{array}
\label{defDresmNet}
\end{equation}
Then we have  
\begin{equation}
\begin{array}{c}
\mathsf{DresNet}(m,\Bell;W)\subset \mathsf{DmresNet}(2^{l_m},m-1,\hat{\Bell};\hat{W}).
\end{array} 
\label{depthReductionResidual}
\end{equation}
 Any function that can be represented by a deep rectifier network with $m$ hidden layers can also be realized by a max-rectifier network with only one hidden layer, more precisely, there exist $\tilde{W}=[\tilde{\mbb}_1,\tilde{W}_1]\in \mathbb{R}^{L\times (l_0+1)}$,  where $l_0$ is the dimension of $\mbx$ and
\begin{equation}
L=\left\{\begin{array}{l}
\dsum_{k=1}^{m/2} 2^{k-1}(l_{2k-1}+l_{2k}),\;\mathrm{if} \; $m$ \;\mathrm{is \; even}\\
2^{\frac{m+1}{2}} l_m + \dsum_{k=1}^{(m-1)/2} 2^{k-1}(l_{2k-1}+l_{2k}),\;\mathrm{if} \; $m$ \;\mathrm{is \; odd}\\
\end{array}\right.
\label{L-formula}
\end{equation} 
 such that
\begin{equation}
\begin{array}{rcl}
\mathsf{DresNet}(m,\Bell;W)&\subset& \mathsf{DmrNet}(2^N,1,L;\tilde{W})
\end{array} 
\label{Dres2Br}
\end{equation}
where 
\begin{small}
\begin{equation}
\begin{array}{rcl}
N&=&\left\{\begin{array}{l}\dsum_{k=2}^{\frac{m}{2}}\mu(k)l_{2k-1} + \dsum_{k=1}^{\frac{m}{2}}\{\mu(k)+1\} l_{2k},\; \mathrm{if}\; m\; \mathrm{is\; even};\\  
\dsum_{k=2}^{\frac{m+1}{2}}\mu(k)l_{2k-1} + \dsum_{k=1}^{\frac{m-1}{2}}\{\mu(k)+1\} l_{2k},\; \mathrm{if}\; m\; \mathrm{is\; odd}
\end{array}\right.\\
\mu(k)&\triangleq& 3(2^{k-1}-1).
\end{array}
\label{N-formula}
\end{equation}
\end{small}
\end{thm}

{\bf Proof}: Note that $\mathsf{DresNet}(m,\Bell,W)=\mathsf{DrNet}(m,\Bell,W)$ when $m\leq 2$. from Theorem \ref{Thm_DrNet}, we have
\begin{equation}
\mathsf{DresNet}(2,\Bell)\subset \mathsf{DmrNet}(2^{l_2},1,l_1+l_2)
\end{equation}
that is, (\ref{Dres2Br}) holds with $L$ and $N$ being defined as in (\ref{L-formula}) and (\ref{N-formula}) respectively.

When $m=3$, from Theorem 6, we have 
\begin{equation}
\begin{array}{rcl}
\mathsf{DresNet}(3,\Bell)&\subset& \mathsf{DmresNet}(2^{l_3},2,[l_1,l_2+2l_3]^T)\\
&\subset& \mathsf{DmrNet}(2^{l_2+3l_3},1,l_1+l_2+2l_3)
\end{array}
\end{equation}
which implies that (\ref{Dres2Br}) holds with $L$ and $N$ being defined as in (\ref{L-formula}) and (\ref{N-formula}) respectively.

We have proved Theorem 7 when $m=2$ and $m=3$. Now assume that (\ref{Dres2Br}) holds when $m=j$ and $j$ is odd.

 Apply Theorem 6 for $m=j+1$ and note that $j+1$ is even, we have 
\begin{equation}
\begin{array}{rcl}
\mathsf{DresNet}(j+1,\Bell)&\subset& \mathsf{DmresNet}(2^{l_{j+1}},j,\hat{\Bell})\\
\hat{\Bell}&\triangleq& [l_1,\cdots,l_{m-2},l_{m-1}+l_m]^T.\\
\end{array}
\label{ReductionJ+1toJ}
\end{equation}

Since Theorem 7 is assumed to be true when $m=j$, and note that $j$ is odd and $\hat{l}_j=l_j+l_{j+1}$, we have 
\begin{equation}
\begin{array}{rcl}
\mathsf{DresNet}(j,\hat{\Bell})&\subset& \mathsf{DmrNet}(2^{\hat{N}},1,L)
\end{array}
\end{equation}
where
\begin{equation}
\begin{array}{rcl}
L&=& 2^{\frac{j+1}{2}} \hat{l}_j + \dsum_{k=1}^{(j-1)/2} 2^{k-1}(l_{2k-1}+l_{2k})\\
&=& 2^{\frac{j+1}{2}} (l_j+l_{j+1}) +\dsum_{k=1}^{(j-1)/2} 2^{k-1}(l_{2k-1}+l_{2k})\\
&=& \dsum_{k=1}^{(j+1)/2} 2^{k-1}(l_{2k-1}+l_{2k})\\
\hat{N}&=&\frac{j+1}{2}\mu\left(\frac{j+1}{2}\right)\hat{l}_j + \dsum_{k=2}^{\frac{j-1}{2}}\mu(k)l_{2k-1} \\
&& + \dsum_{k=1}^{\frac{j-1}{2}}\{\mu(k)+1\} l_{2k}\\
&=&\frac{j+1}{2}\mu\left(\frac{j+1}{2}\right)(l_j+l_{j+1}) + \dsum_{k=2}^{\frac{j-1}{2}}\mu(k)l_{2k-1} \\
&& + \dsum_{k=1}^{\frac{j-1}{2}}\{\mu(k)+1\} l_{2k}\\
&=&-l_{j+1} + \dsum_{k=2}^{\frac{j+1}{2}}\mu(k)l_{2k-1} \\
&& + \dsum_{k=1}^{\frac{j+1}{2}}\{\mu(k)+1\} l_{2k}.
\end{array} 
\end{equation}
Therefore
\begin{equation}
\begin{array}{rcl}
\mathsf{DmresNet}(2^{l_{j+1}},j,\hat{\Bell})&\subset& \mathsf{DmresNet}(2^N,1,L)\\
\end{array}
\label{resTemp1}
\end{equation}
where
\begin{equation}
\begin{array}{rcl}
N&=& \hat{N}+l_{j+1}\\
&=& \dsum_{k=2}^{\frac{j+1}{2}}\mu(k)l_{2k-1} + \dsum_{k=1}^{\frac{j+1}{2}}\{\mu(k)+1\} l_{2k}.
\end{array}
\end{equation}
From (\ref{resTemp1}) and (\ref{ReductionJ+1toJ}), we know that Theorem 7 holds when $m=j+1$ for the case with $j$ being odd. 

Similarly, one can prove that Theorem 7  holds when $m=j+1$ for the case with $j$ being even.  Then by mathematical reduction, Theorem 7 holds for any $m\geq 2$. 

\begin{flushright}
$\Box$
\end{flushright}

\section{Comparisons of Plain Nets and Residual Nets}

To compare the plain nets and residual nets, we assume that both nets have the same depth, denoted by $m$, and have the same total number, denoted by $T$, of hidden units.  We also assume that each hidden layer has the same number of hidden nodes, that is, $l_i=l$ for all $i$ but the width $l$ is different for the plain net and the residual net since the residual net has skip connections. For plain nets, the width is $\frac{T}{m}$. For residual nets, the width is $\frac{2T}{3m}$. Then from Theorem 1 and Theorem 7, we have
\begin{equation}
\begin{array}{rcl}
L_p&=& T\\
N_p &=& \frac{m-1}{2}T\\
L_{res}&=& T\left\{2^{0.5m+1}-\frac{m(m+2)}{4}\right\}\\
N_{res} &=& \frac{2T}{3m}\left\{2^{0.5m+1}-2\right\}
\end{array}
\label{comparison}
\end{equation}
where $L_p,2^{N_p}$ are respectively the maximum number of hidden nodes and the maximum number of linear units in the output layer of the single hidden layer max-rectifier nets transformed from the plain nets, and $L_{res},2^{N_{res}}$ are those of the transformed shallow nets from the residual nets.

From (\ref{comparison}), one can conclude that: 1) The number of hidden nodes increases linearly with the total number of hidden units; 2) The number of hidden nodes remains constant with the depth of plain nets, but increases exponentially with the depth of residual nets; 3) the number of the linear units in the output layer increases exponentially with the depth of plain nets but more than exponentially with the depth of residual nets.

\section{Conclusion}

In this paper, we have developed transformations capable of converting deep rectifier neural nets into shallow rectifier nets and used them to analyse different types of learning architectures. From these transformations, one can appreciate the superior expressive power of deep nets compared to shallow nets, and the advantages of adding skip connections in deep residual networks.



\begin{small}
\bibliography{icml2017bib}
\bibliographystyle{icml2017}
\end{small}

\end{document}